\documentclass[10pt,twocolumn,letterpaper]{article}

\usepackage{iccv}
\usepackage{times}
\usepackage{epsfig}
\usepackage{graphicx}
\usepackage{amsmath}
\usepackage{amssymb}

\usepackage{booktabs}
\usepackage{multirow}
\usepackage{makecell}
\usepackage{amsfonts}
\usepackage{comment}
\usepackage{xcolor}
\usepackage{epstopdf}

\usepackage{multicol}
\usepackage{diagbox}
\usepackage{wrapfig}
\usepackage{picinpar}
\usepackage{subfloat}
\usepackage{blindtext}
\usepackage{caption}
\usepackage{floatrow}
\newfloatcommand{capbtabbox}{table}[][\FBwidth]
\usepackage{comment}
\newcommand{\fengx}[1]{\textcolor[rgb]{0,0,1} {#1}}
\newcommand{\pei}[1]{\textcolor{red}{#1}}
\usepackage{enumitem}

\usepackage[accsupp]{axessibility}

\usepackage[pagebackref=true,breaklinks=true,letterpaper=true,colorlinks,bookmarks=false]{hyperref}

\iccvfinalcopy 

\def\iccvPaperID{6339} 

\ificcvfinal\fi

\begin{document}

\title{Hierarchical Contrastive Learning for Pattern-Generalizable Image \\Corruption Detection}

\author{Xin Feng\footnotemark[1]\qquad Yifeng Xu\footnotemark[1]\qquad Guangming Lu\footnotemark[2]\qquad Wenjie Pei\footnotemark[2]\\
Harbin Institute of Technology, Shenzhen
}

\twocolumn[{%
\renewcommand\twocolumn[1][]{#1}%
\maketitle
\ificcvfinal\thispagestyle{empty}\fi
\vspace{-32pt}
\begin{center}
    \includegraphics[width=1\textwidth]{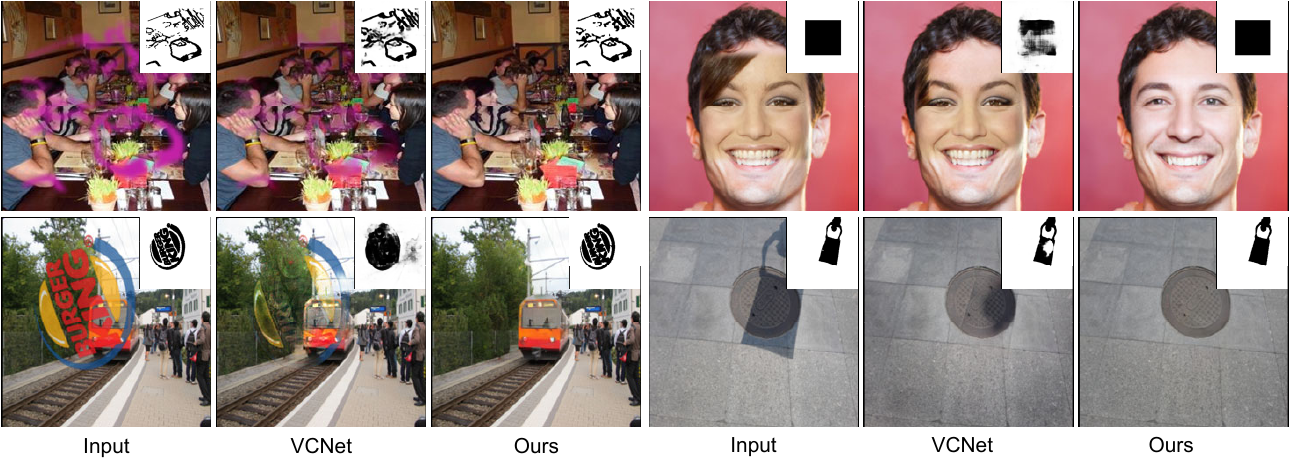}
    \vspace{-20pt}
    \captionof{figure}{Visual comparison between our method and VCNet~\cite{wang2020vcnet} (the state-of-the-art method for blind image inpainting) on four tasks: blind image inpainting (graffiti corruption) (\textbf{up left}), blind image inpainting (real-image-patch corruption) (\textbf{up right}), watermark removal (\textbf{bottom left}) and shadow removal (\textbf{bottom right}). The groundtruth mask and the predicted masks by different methods are presented.} 
    \label{fig:teaser}
\end{center}
\vspace{-2pt}
}]

\renewcommand{\thefootnote}{\fnsymbol{footnote}} 
\footnotetext[1]{These authors contributed equally to this work.}
\footnotetext[2]{Corresponding authors.} 

\begin{abstract}
\vspace{-8pt}

Effective image restoration with large-size corruptions, such as blind image inpainting, entails precise detection of corruption region masks which remains extremely challenging due to diverse shapes and patterns of corruptions. In this work, we present a novel method for automatic corruption detection, which allows for blind corruption restoration without known corruption masks. Specifically, we develop a hierarchical contrastive learning framework to detect corrupted regions by capturing the intrinsic semantic distinctions between corrupted and uncorrupted regions. In particular, our model detects the corrupted mask in a coarse-to-fine manner by first predicting a coarse mask by contrastive learning in low-resolution feature space and then refines the uncertain area of the mask by high-resolution contrastive learning. A specialized hierarchical interaction mechanism is designed to facilitate the knowledge propagation of contrastive learning in different scales, boosting the modeling performance substantially. 
The detected multi-scale corruption masks are then leveraged to guide the corruption restoration. Detecting corrupted regions by learning the contrastive distinctions rather than the semantic patterns of corruptions, our model has well generalization ability across different corruption patterns. Extensive experiments demonstrate following merits of our model: 1) the superior performance over other methods on both corruption detection and various image restoration tasks including blind inpainting and watermark removal, and 2) strong generalization across different corruption patterns such as graffiti, random noise or other image content. Codes and trained weights are available at \url{https://github.com/xyfJASON/HCL}.
\vspace{-8pt}

\end{abstract}

\vspace{-8pt}
\section{Introduction}
\label{sec:intro}
\vspace{-4pt}

An essential yet challenging step for image restoration with large-size corruptions, such as blind image inpainting, watermark removal or shadow removal, is to detect the corruption region masks in pixel level precisely. The difficulties lie in the diverse shapes and appearance of corruptions, which can be hardly modeled in a uniform pattern. As a result, blind image inpainting remains a challenging task although image inpainting with known corruption masks~\cite{li2022mat, dong2022incremental,feng2022generative,liu2022reduce} has achieved remarkable progress.

A straightforward way for corruption detection is to formulate it as a segmentation task and predict the corruption mask by pixel-wise binary classification. A prominent example following such modeling paradigm is VCNet~\cite{wang2020vcnet}, a state-of-the-art method for blind image inpainting, which regards corruptions as target objects and learns to recognize the semantic patterns of corruptions for detection. While VCNet can successfully detect the corruptions with uniform patterns, it has two potential limitations. First, the corruption may exhibit diverse patterns due to its potentially irregular appearance nature, which VCNet can hardly deal with. Second, it is challenging for VCNet to handle the corruption patterns that are distinct from those appearing in training data, thus it has limited generalization w.r.t. different corruption patterns. 



In this work, we propose a novel model for image corruption detection to circumvent the aforementioned limitations of VCNet. Instead of recognizing the semantic patterns of corruptions as VCNet does, we apply metric learning to learn an embedding space in which our model can capture the contrastive semantic distinctions between the corrupted and uncorrupted regions. 
To this end, we design a hierarchical contrastive learning framework for detecting corrupted regions, which predicts the corruption mask in a coarse-to-fine manner. To be specific, it first predicts a coarse mask by lightweight contrastive learning in a low-resolution feature space. Then it refines the uncertain pixels with low confidence in the predicted mask by high-resolution contrastive learning based on fine-grained features. Note that only a small fraction of predicted masks need to be refined, thus the refining process can be performed quite efficiently. To facilitate the knowledge propagation and consistency of contrastive learning between different scales, we propose a specialized hierarchical interaction mechanism. As a result, the high-resolution contrastive learning can inherit useful knowledge from the low-resolution contrastive learning to achieve precise prediction of corruption masks quite efficiently. The detected multi-scale corruption masks are further used to guide the restoration of the corrupted regions, which also follows a coarse-to-fine generative process. 

Unlike VCNet recognizing the semantic patterns of corruptions, our model focuses on capturing the contrastive semantic distinctions between corrupted and uncorrupted regions. Thus our model has well generalization ability across different corruption patterns, As shown in Figure~\ref{fig:teaser}. To conclude, we make following contributions. 
\begin{itemize}[leftmargin =*, itemsep = -4pt, topsep = -2pt]
\item We design a hierarchical contrastive learning framework to detect multi-scale corruption masks by learning the contrastive semantic distinctions between corrupted and uncorrupted regions. 
\item Integrating the proposed hierarchical contrastive learning module, we develop a general-purpose blind image restoration model to perform high-quality image restoration without known corruption masks.
\item Extensive experiments validate the effectiveness of our model both quantitatively and qualitatively in three aspects: 1) superior performance on corruption detection (Section~\ref{sec:blind}), 2) favorable performance compared to the specialized methods on two challenging image restoration tasks including blind image inpainting and watermark removal (Section~\ref{sec:adaption}), and 3) strong generalizability across different corruption patterns (Section~\ref{sec:blind} and~\ref{sec:general}). 
\end{itemize}

\vspace{-4pt}
\section{Related Work}
\label{sec:relatedwork}

\noindent\textbf{Image Restoration with known Corruption masks.}
Image restoration with given corruption masks assumes that the corrupted regions are known, and conventional methods~\cite{barnes2009patchmatch,efros2001image,criminisi2004region} leverage uncorrupted content to restore the corrupted regions by pixel diffusion or patch replacement. 
However, these methods usually generate significant artifacts in the restored image due to the lack of global semantic understanding and generalization ability. With the great success of convolutional neural networks(CNNs) in computer vision, recent methods leverage an encoder-decoder framework to restore corrupted images.
A typical way of CNN-based methods~\cite{liu2018image,yu2019free,Li_2020_CVPR,li2022mat} employs known uncorrupted regions to infer corrupted content from the outside to the inside.
To synthesize reasonable semantics, some methods~\cite{liu2019coherent,Nazeri_2019_ICCV,jie2020inpainting,Liu2019MEDFE,Guo_2021_ICCV,li2022misf} leverage structure information to improve the quality of semantic structures in the corrupted regions.
Promoted by generative models~\cite{NIPS2014_5ca3e9b1,kingma2013auto},
some recent methods~\cite{dong2022incremental,liu2022reduce,zheng2019pluralistic} attempt to learn generative prior for improving the quality of synthesized images, and generating diverse content for corrupted regions.

\begin{figure*}[!t]
    \centering
    \begin{minipage}[b]{0.85\linewidth}
    \includegraphics[width=\linewidth]{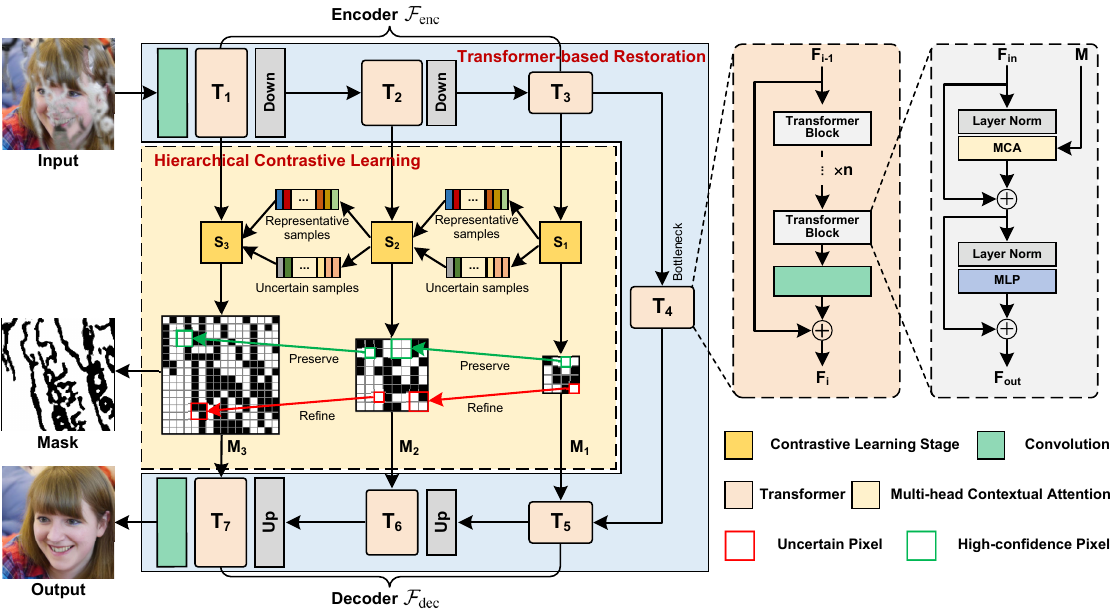}
    \end{minipage}
    \caption{Architecture of our method. It leverages the designed Hierarchical Contrastive Learning to detect corruption masks and further synthesizes reasonable content for corrupted regions by the developed Transformer-based Restoration module.
    \vspace{-12pt}
} 
    \label{fig3:framework}
    \vspace{-8pt}
\end{figure*}

\noindent\textbf{Image corruption detection.}
Early methods~\cite{cai2017blind,liu2019deep} for image corruption detection assume the corrupted content follows a simple distribution, such as constant values, Gaussian noise, etc. However, previous assumption simplifies the detection of corrupted regions, limiting the application scope of image corruption detection. Recently, Wang \emph{et al.}~\cite{wang2020vcnet} relax the definition of image corruption detection by increasing the diversity of corruption types, including watermarking, raindrop, or even random images. They also propose a two-stage framework \emph{VCNet} for blind image restoration, which first locates corrupted regions in a manner of image segmentation and then fills in content following the typical encoder-decoder based methods for image restoration. However, it sometimes misuses content in corrupted regions while reconstructing the complete image due to the challenge of detecting diverse patterns of corruption.

\noindent\textbf{Contrastive learning.}
Contrastive learning has been widely used in self-supervised representation learning \cite{doersch2015unsupervised, he2020momentum, chen2020simple}. Instead of matching an input image to a fixed target, contrastive learning maximizes mutual information in the representation space, keeping the query close to the positive sample and away from the negative sample. Previous works \cite{he2020momentum, chen2020simple,wang2021dense} have validated the effectiveness of contrastive learning in high-level vision tasks, due to the inherent suitability for modeling feature contrasts. 
Recently, some researchers~\cite{wang2021unsupervised,wu2021contrastive} have attempted to introduce contrastive learning into low-level vision tasks. However, pixel-wise contrastive learning on high-resolution images suffers from large computational cost and limited performance. In this paper, we design a hierarchical contrastive learning mechanism to solve above problems and significantly strengthen the performance of image corruption detection.


\vspace{-8pt}
\section{Approach}
\label{sec:method}
\vspace{-4pt}
Detecting image corruptions by recognizing the semantic patterns of corruptions is difficult in that the corruptions can potentially exhibit diverse patterns. Our proposed hierarchical contrastive learning framework learns the contrastive semantic distinctions between corrupted and uncorrupted regions, and detects corruption masks in a coarse-to-fine manner. The obtained multi-scale corruption masks are then leveraged to guide the restoration of the corruptions in a coarse-to-fine generative process.

\vspace{-4pt}
\subsection{Overview}
\vspace{-4pt}
\noindent\textbf{Problem formulation.}
Given an input image $\textbf{I}\in\mathbb{R}^{C\times H\times W}$ which is corrupted from an intact groundtruth image $\textbf{O}_{\text{gt}}\in\mathbb{R}^{C\times H\times W}$, the task of blind image restoration aims to first detect the corruption region mask $\textbf{M}\in\mathbb{R}^{1\times H\times W}$ and then reconstruct a complete image $\hat{\textbf{O}}\in\mathbb{R}^{C\times H\times W}$ by synthesizing realistic content $\hat{\textbf{C}}$ for the corrupted regions:
\vspace{-8pt}
\begin{small}
\begin{equation}
\begin{split}
    \label{equ:def}
    &\textbf{I} = \textbf{O}_{\text{gt}}\odot \textbf{M} + \textbf{N}\odot(1-\textbf{M}), \\
    &\hat{\textbf{O}} = \textbf{O}_{\text{gt}}\odot \textbf{M} + \hat{\textbf{C}}\odot(1-\textbf{M}).
\end{split}
\vspace{-4pt}
\end{equation}    
\end{small}

\noindent\text{Herein}, $\textbf{M}\in\mathbb{R}^{1\times H\times W}$ is a binary map where the values of corrupted regions are 0 and uncorrupted regions are filled with 1. $\textbf{N}\in\mathbb{R}^{C\times H\times W}$ is the noisy content in the corrupted regions. Precise detection of the corruption mask $\textbf{M}$ is crucial to the performance of image restoration.

As shown in Figure~\ref{fig3:framework}, our model follows the encoder-decoder framework and consists of two core modules: Hierarchical Contrastive Learning module for detecting corruption masks and Transformer-based Restoration module for corruption restoration. The corrupted image is first encoded into multi-scale feature maps by Encoder $\mathcal{F}_\text{enc}$ of Transformer-based Synthesis module. Then the proposed Hierarchical Contrastive Learning module is employed to perform corruption detection in a coarse-to-fine manner from these feature maps, and predicts multi-scale corruption masks. Finally, the obtained masks are fed into Decoder $\mathcal{F}_\text{Dec}$ of the Transformer-based Synthesis module for restoration, which is also in a coarse-to-fine generative process. 
Both Encoder $\mathcal{F}_\text{enc}$ and Decoder $\mathcal{F}_\text{dec}$ are mainly composed of basic transformer blocks of MAT~\cite{vaswani2017attention}, while they are equipped with additional down-sampling layers and up-sampling layers respectively. We employ a shallow convolutional layer with $5\times5$ kernel before Encoder and after Decoder to perform basic feature transformation. Besides, we also employ another Conv-U-Net to refine high-frequency details of output results, leaning upon the local texture refinement capability and efficiency of CNNs.

\vspace{-4pt}
\subsection{Corruption Detection by Hierarchical Contrastive Learning}
\vspace{-4pt}

We design Hierarchical Contrastive Learning framework to guide the learning of multi-scale semantic embedding spaces of Encoder $\mathcal{F}_\text{enc}$ in such a way that the semantic distances between two arbitrary pixels both within uncorrupted or corrupted regions (intra-region distance) should be minimized while the distance between a corrupted pixel and an uncorrupted pixel (inter-region distance) should be maximized. As a result, our model can capture the intrinsic semantic distinctions between corrupted and uncorrupted regions. Then our model performs clustering on all pixels in this learned embedding space to separate them into two clusters, corresponding to the corrupted and uncorrupted regions, and thus achieves the corruption mask.

As shown in Figure~\ref{fig3:framework}, Encoder $\mathcal{F}_\text{enc}$ consists of three encoding stages and produces three scales of encoded feature maps. Accordingly, Our model performs three stages of pixel-level contrastive learning for the corresponding scale of feature maps to guide the learning of its embedding space. We will first describe how our model performs corruption detection in one stage with single-scale contrastive learning. Then we will elaborate on the proposed Hierarchical Interaction Mechanism which enables our model to perform hierarchical contrastive learning quite efficiently for coarse-to-fine multi-scale mask detection.

\vspace{-6pt}
\subsubsection{Single-Scale Contrastive Learning}
\noindent\textbf{Supervised Contrastive Learning.} During each stage of contrastive learning, we construct positive pixel pairs by using intra-region pixels, i.e., both pixels are from either the uncorrupted region or the corrupted region. In contrast, each negative training pair consists of two inter-region pixels, one from the corrupted region and the other from the uncorrupted region. To be specific, for a query pixel $q$ from a randomly selected query set $Q$, we construct the positive set $P$ by randomly selecting pixels from the same region as $q$ and construct the negative set $N$ from the opposite region. Then we apply Circle loss~\cite{sun2020circle} to maximize the cosine similarity of positive pairs while minimizing the similarities of negative pairs. Formally, the contrastive learning in the stage $\text{S}_s$ is guided by the loss: 
\vspace{-4pt}
\begin{equation}
\resizebox{0.9\linewidth}{!}{$
    \mathcal{L}^s_\text{CL}\!=\!\!\sum_{q \in Q} \!\text{log}\!\left[1+\!\sum\limits_{p\in P}\!\text{exp}(-\mathbf{e}_{q}\!\cdot\!\mathbf{e}_{p}\!/\tau)\! \cdot \!\sum\limits_{n\in N}\!\text{exp}(\mathbf{e}_{q}\!\cdot\!\mathbf{e}_{n}\!/\tau)\!\right]\!,\!
    $}
    \label{eqn:cl}
    \vspace{-4pt}
\end{equation}
where $\tau$ is a scale factor. $\mathbf{e}_q$ is the features for pixel $q$ projected from the corresponding embedding space of Encoder $\mathcal{F}^s_\text{enc}$ by a projection head in Stage-$s$:
\vspace{-4pt}
\begin{equation}
    \mathbf{e}_q = \mathcal{F}^s_{\text{proj}}(\mathcal{F}^s_\text{enc}(\mathbf{I}_q)),
    \label{eqn:proj}
    \vspace{-4pt}
\end{equation}
where the projection head $\mathcal{F}_\text{proj}$ comprises two fully connected layers and a \emph{GELU}~\cite{hendrycks2016gaussian} layer in-between for non-linear transformation.

\noindent\textbf{Corruption Mask Detection.} Under the supervision of contrastive learning in Equation~\ref{eqn:cl}, the pixels within the same region, either the corrupted or the uncorrupted regions, tend to have similar representations in each scale of embedding space of Encoder $\mathcal{F}_\text{enc}$ while the pixels from different regions would have dissimilar representations. Thus, we can perform clustering to separate them into two clusters. Specifically, our model adopts K-means algorithm for clustering, and the clustering in Stage-$s$ is performed by:
\vspace{-4pt}
\begin{small}
\begin{equation}
[\mathbf{c}^s_1, \mathbf{c}^s_2], [\mathbf{M}^s_1, \mathbf{M}^s_2] = \text{K-means}(\mathcal{F}^s_{\text{proj}}(\mathcal{F}^s_\text{enc}(\mathbf{I}))),
\vspace{-4pt}
\end{equation}    
\end{small}

\noindent\text{where} $[\mathbf{c}^s_1, \mathbf{c}^s_2]$ denotes the embeddings of two produced cluster centers and $[\mathbf{M}^s_1, \mathbf{M}^s_2]$ are the associated binary masks. To identify which cluster corresponds to the corrupted or uncorrupted regions, we train a lightweight binary classifier consisting of two fully connected layers with a \emph{ReLU} layer, and use it to perform classification on two cluster centers. Thus, the associated mask for the cluster of the corrupted region is the predicted corruption mask.

\begin{figure}[!t]
    \centering
    \begin{minipage}[b]{1\linewidth}
    \includegraphics[width=\linewidth]{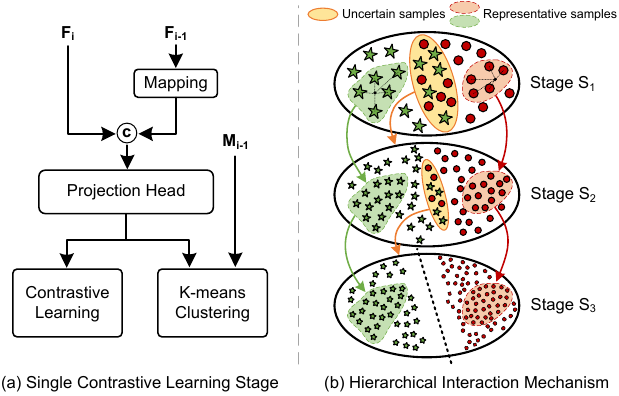}
    \end{minipage}
    \vspace{-10pt}
    \caption{In the stages $\text{S}_2$ and $\text{S}_3$, both the selected representative samples and uncertain samples in the previous stage are used for contrastive learning in current stage while only the mask labels of uncertain samples are refined during mask detection.
    \vspace{-12pt}}
    \label{fig4:hcl}
\end{figure}

\vspace{-8pt}
\subsubsection{Hierarchical Interaction Mechanism}
\vspace{-3pt}
Our model performs Hierarchical Contrastive Learning including three stages to detect the corruption mask in a coarse-to-fine manner. 
It first performs contrastive learning in the lowest-resolution stage ($\text{S}_1$) to predict a coarse mask for corruption. Then it refines the uncertain pixels of the mask with low confidence by higher-resolution contrastive learning in subsequent stages ($\text{S}_2$ and $\text{S}_3$). We propose Hierarchical Interaction Mechanism to facilitate the interaction and knowledge propagation between different stages of contrastive learning and guarantee the semantic consistency between them. In particular, higher-resolution contrastive learning can inherit useful knowledge from previous stages of contrastive learning, which can improve the learning performance substantially. 
Quadtree structure are used to build the positional correspondence between feature maps in adjacent stages considering that the resolution of feature maps is always scaled by four times between adjacent stages.

\noindent\textbf{Selecting High-Quality Training Samples.}
During the contrastive learning in $\text{S}_1$, we construct the query set $Q$ as well as the associated positive set $P$ and negative set $N$ in Equation~\ref{eqn:cl} by randomly selecting pixels from the whole feature map. To improve the efficiency of contrastive learning in the higher stages ($\text{S}_2$ and $\text{S}_3$), we select a small fraction of pixels from the entire feature maps, which are crucial for contrastive learning, as the high-quality candidate set for constructing $Q$, $P$ and $N$.

To be specific, we first measure the prediction confidence of each pixel based on the clustering results of last stage. The prediction confidence of the pixel $q$ in the stage $\text{S}_s$ is calculated by:
\vspace{-6pt}
\begin{equation}
    z^s_q = \frac{\text{exp}({-(\mathbf{e}_q \cdot \mathbf{c}_{y_q}^s/\tau}))}{\sum_{i=1}^2 \text{exp}({-(\mathbf{e}_q \cdot \mathbf{c}_{i}^s/\tau}))},
    \vspace{-4pt}
    \label{eqn:confidence}
\end{equation}
where $y_q$ denotes the cluster index $q$ belongs to and $\tau$ is the scaling factor. Then we pick out two types of pixels as the high-quality query set for the stage $\text{S}_{s+1}$: 1) the pixels with high confidence which are regarded as representatives of two clusters and are typically close to the cluster centers; 2) the uncertain pixels with low confidence which are mostly near the boundaries between the corrupted and uncorrupted regions. Selecting pixels in such a way for constructing the positive and negative training pairs, our model is able to perform high-resolution contrastive learning in the stages $\text{S}_2$ and $\text{S}_3$ quite efficiently and effectively. 

\noindent\textbf{Inter-Stage Semantic Consistency.}
To guarantee the semantic consistency between different stages of contrastive learning, we reuse the learned features of the previous stage of contrastive learning in current stage. As shown in Figure~\ref{fig4:hcl}, we concatenate the features in the previous stage with the features in the current stage, which are then fed into the projection head to produce input features for current stage. Accordingly, Equation~\ref{eqn:proj} for high-resolution stages ($\text{S}_2$ and $\text{S}_3$) evolves into the following form:
\vspace{-4pt}
\begin{equation}
    \mathbf{e}_q = \mathcal{F}^s_{\text{proj}}(\text{Concat}(\mathcal{F}^s_\text{enc}(\mathbf{I}_q), \mathcal{F}^s_\text{map}(\mathcal{F}^{s-1}_\text{enc}(\mathbf{I}_q)))),
    \vspace{-4pt}
    \label{eqn:proj2}
\end{equation}
\noindent\text{where} the mapping function $\mathcal{F}^s_\text{map}$ is a linear transformation to regulate the feature dimension of previous stage and balance the effect between current and previous features. Besides the semantic consistency between adjacent stages, another merit of such feature reusing is that our model is able to inherit the learned semantics from previous stage.

\noindent\textbf{Refining the Mask Prediction of Uncertain Pixels.} 
During the high-resolution contrastive learning in the stages $\text{S}_2$ and $\text{S}_3$, we only refine the mask prediction by re-predicting the labels of uncertain pixels with low confidence in the previous stage, i.e., the type-2 samples of selected high-quality query set. The mask labels of other pixels with high confidence are directly inherited from the predictions in the previous stage according to the built quadtree, assuming that the predictions of these pixels are reliable. Thus, such refining process can be performed quite efficiently.

\vspace{-2pt}
\subsection{Transformer-based Restoration Module}
As shown in Figure~\ref{fig3:framework}, the detected corruption masks are fed into Decoder of the Transformer-based Restoration module for corruption restoration, which consists of the designed mask-guided transformer block.

\noindent\textbf{Mask-guided transformer block.}
As illustrated in Figure~\ref{fig3:framework}, our transformer block contains a multi-head contextual attention (MCA) module, followed by a two-layer MLP with \emph{GELU} nonlinearity in between. Besides, a layer normalization (LN)~\cite{ba2016layer} layer is applied before each attention and each MLP, and a residual learning connection is employed after each module:
\vspace{-4pt}
\begin{equation}
\begin{split}
    F'^s_i &= \text{MCA}(\text{LN}(F^s_{i-1}), \text{M}^s)+F^s_{i-1} , \\
    F^s_i  &= \text{MLP}(\text{LN}(F'^s_i) ) + F'^s_i,
\end{split}
\end{equation}
where $F^s_i$ is the output features of the $i$-th block in the stage $\text{S}_s$. 
To fully exploit the predicted masks by hierarchical contrastive learning, we employ the multi-head contextual attention module proposed in MAT~\cite{li2022mat},
which follows the shifted window manner~\cite{liu2021swin} and is formulated as:
\vspace{-4pt}
\begin{small}
\begin{equation}
    \text{Att}(\textbf{Q},\textbf{K},\textbf{V}) = \text{Softmax}(\frac{\textbf{Q}\textbf{K}^\text{T}+\hat{\mathbf{M}}^s}{\sqrt{d_k}})\textbf{V},
    \vspace{-4pt}
\end{equation}
\end{small}

\noindent\text{where} $\textbf{Q}, \textbf{K}, \textbf{V}$ are the query, key, and value matrices, respectively. $\sqrt{d_k}$ is the scaling factor. $\hat{\mathbf{M}}$ is defined by
\vspace{-4pt}
\begin{equation}
    \hat{\mathbf{M}}^s_i = \gamma(\mathbf{M}^s_i-1),
    \vspace{-4pt}
\end{equation}
where $\gamma$ is a large positive integer to reduce the values of corrupted pixels.


\begin{figure*}[!t]
    \centering
    \begin{minipage}[b]{0.9\linewidth}
    \includegraphics[width=\linewidth]{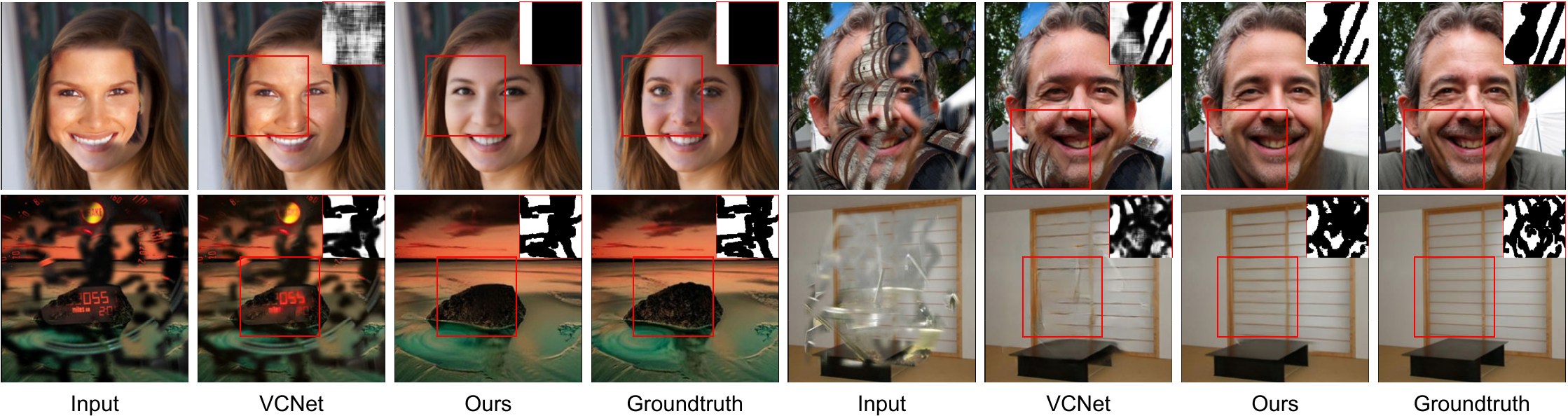}
    \end{minipage}
    \caption{Visualization of detected masks and reconstructed images on four randomly selected samples from test set.
    \vspace{-12pt}} 
    \label{fig10:results}
    \vspace{-8pt}
\end{figure*}
\vspace{-2pt}
\subsection{Joint Optimization for Parameter Learning}
The proposed Hierarchical Contrastive Learning module and the Restoration module are integrated in a holistic way based on the encoding-decoding framework, forming a general-purpose blind image restoration model, as illustrated in Figure~\ref{fig3:framework}. The whole model is optimized jointly. Besides the loss of hierarchical contrastive learning (Equation~\ref{eqn:cl}), we also adopt three more supervision signals for corruption restoration, including pixel-wise reconstruction loss, perceptual loss, and adversarial loss.

\noindent\textbf{Pixel Reconstruction Loss}, which pushes the restored image $\hat{\mathbf{O}}$ as close as its groundtruth $\mathbf{O}_{\text{gt}}$:
\vspace{-4pt}
\begin{equation}
    \mathcal{L}_{\text{pixel}}=\left\|{\hat{\mathbf{O}}}-\mathbf{O}_{\text{gt}}\right\|_1.
    \vspace{-6pt}
\end{equation}

\noindent\textbf{Perceptual Loss}~\cite{johnson2016perceptual}, which performs semantic supervision on the restored images in the deep feature space: 
\vspace{-8pt}
\begin{small}
\begin{equation}
    \mathcal{L}_{\text{perc}} = \sum_{l=1}^L \frac{1}{C_l  H_l W_l} \left\| f^l_\text{vgg}(\hat{\mathbf{O}}) - f^l_\text{vgg}(\mathbf{O}_{\text{gt}})\right\|_1.
    \vspace{-6pt}
\end{equation}
\end{small}

\noindent\text{Herein} $f^l_\text{vgg}(\hat{\mathbf{O}})$ and $f^l_\text{vgg}(\mathbf{O}_{\text{gt}})$ are the extracted features from $\hat{\mathbf{O}}$ and $\mathbf{O}_{\text{gt}}$ respectively from the $l$-th convolution layer of the pre-trained VGG-19 network~\cite{simonyan2014very}.

\noindent\textbf{Adversarial Loss}, which employs WGAN-GP~\cite{gulrajani2017improved} to encourage $\hat{\mathbf{O}}$ to be as realistic as its groundtruth $\mathbf{O}_{\text{gt}}$:
\vspace{-6pt}
\begin{equation}
    \mathcal{L}_{\text{adv}}  = -\mathbb{E}_{\hat{\mathbf{O}}\backsim \mathbb{P}_{\hat{\mathbf{O}}}}\left[D(\hat{\mathbf{O}})\right],
    \vspace{-6pt}
\end{equation}
where $D$ is a discriminator. 
Overall, the whole model is optimized by:
\vspace{-8pt}
\begin{equation}
    \mathcal{L} = \lambda_1\mathcal{L}_{\text{pixel}} + \lambda_2\mathcal{L}_{\text{perc}}+\lambda_3\sum_{s=1}^3\mathcal{L}_{\text{CL}}^s+\lambda_4\mathcal{L}_{\text{adv}},
    \vspace{-6pt}
\end{equation}
where $\lambda_1$, $\lambda_2$, $\lambda_3$, and $\lambda_4$ are hyper-parameters to balance between different losses. 

We integrate the Contrastive Learning module between Encoder and Decoder with consistent coarse-to-fine refining stages for all three modules, which potentially enables synchronized optimization of three modules in each scale.

\section{Experiments}
\label{sec:exp}
We first evaluate the performance of our method on image corruption detection, then validates the effectiveness of our method on two challenging image restoration tasks including blind image inpainting and watermark removal. Finally, we conduct extensive ablation study to obtain more insights into our method.
\subsection{Experimental Settings}
Following the previous work~\cite{wang2020vcnet} for blind image inpainting, we make two pre-processing operations for data generation: 1) we randomly select natural images from large-scale datasets rather than simple constant values or Gaussian noise as the noisy content for corruptions to increase the difficulty of mask detection in blind image inpainting; 2) we smooth the mask boundaries using alpha blending to avoid distinct boundaries between the corrupted and uncorrupted regions.

Four large-scale datasets are used in our experiments, including FFHQ (faces)~\cite{karras2019style}, CelebA-HQ (faces)~\cite{karras2018progressive}, ImageNet (objects)~\cite{deng2009imagenet}, and Places (scenes)~\cite{zhou2017places}. For each dataset, images from different datasets are randomly selected as the noise content for corruptions. For instance, the images from CelebA-HQ and ImageNet are used as noise content for FFHQ. We employ the method~\cite{liu2018image} to generate irregular masks with mask shapes and corruption ratios. We perform experiments on two resolutions of images for comprehensive evaluation: 256$\times$256 and 512$\times$512.

Adam~\cite{kingma2014adam} is used as the optimizer by setting $\beta_1$, $\beta_2$, initial learning rate and batch size to be 0.9, 0.999, 0.0001, and 4, respectively.
More experimental details are provided in the supplementary material.

\begin{table}[!t]
\centering
\vspace{-6pt}
\caption{Performance comparison for mask detection on three benchmark datasets.
}
\resizebox{0.75\linewidth}{!}{
\begin{tabular}{c |l| c c c c c c}
\toprule
\multicolumn{2}{c|}{Dataset}&\multicolumn{2}{c}{FFHQ~\cite{karras2019style}} &\multicolumn{2}{c}{ImageNet~\cite{deng2009imagenet}}&\multicolumn{2}{c}{Places~\cite{zhou2017places}} \\
\cmidrule(lr){1-2}
\cmidrule(lr){3-4} 
\cmidrule(lr){5-6}
\cmidrule(lr){7-8}
\multicolumn{2}{c|}{Mask ratio (\%)}&0-30&30-60&0-30&30-60&0-30&30-60\\
\midrule
\multirow{2}*{Acc~$\uparrow$}&VCNet&0.948&0.943&0.972&0.978&0.974&0.976 \\
~&Ours&\textbf{0.978}&\textbf{0.976}&\textbf{0.982}&\textbf{0.983}&\textbf{0.984}&\textbf{0.985} \\
\midrule
\multirow{2}*{F1~$\uparrow$}&VCNet&0.967&0.948&0.982&0.980&0.984&0.978\\
~&Ours&\textbf{0.986}&\textbf{0.978}&\textbf{0.989}&\textbf{0.985}&\textbf{0.991}&\textbf{0.987} \\
\midrule
\multirow{2}*{BCE~$\downarrow$}&VCNet&0.126&0.137&\textbf{0.073}&\textbf{0.055}&\textbf{0.064}&\textbf{0.060} \\
~&Ours&\textbf{0.090}&\textbf{0.095}&0.083&0.075&0.068&0.064 \\
\midrule
\multirow{2}*{IoU~$\uparrow$}&VCNet&0.931&0.900&0.966&0.962&0.969&0.959 \\
~&Ours&\textbf{0.975}&\textbf{0.959}&\textbf{0.978}&\textbf{0.965}&\textbf{0.982}&\textbf{0.974} \\
\bottomrule
\end{tabular}
 }
\label{tab1:mask}
\vspace{-14pt}
\end{table} 
\vspace{-4pt}
\subsection{Image Corruption Detection}
\label{sec:blind}
\vspace{-2pt}
\noindent\textbf{Accuracy of corruption detection.} We first compare the accuracy of corruption detection between our model and VCNet, the state-of-the-art method for corruption detection in blind image inpainting. Four metrics are used for comprehensive evaluation: binary cross entropy (BCE), classification accuracy (Acc), F1 score, and intersection over union (IoU). Table~\ref{tab1:mask} lists the results for different corruption ratios on three datasets in 256$\times$256 resolution, which show that our model achieves superior performance of mask detection over VCNet in terms of all metrics except BCE. This is reasonable because VCNet is directly supervised by the BCE loss. Besides, we also make qualitative comparison in Figure~\ref{fig10:results}, in which our model detects more precise corruption masks and restores higher-quality images than VCNet.

\noindent\textbf{Image inpainting based on detected corruption masks by different methods.} As an indirect evaluation of corruption detection, we perform image inpainting over the detected corruption masks from different methods, employing the same inpainting model MAT~\cite{li2022mat}, and then compare the inpainting performance. To have a comprehensive evaluation, we also provide the inpainting performance over the detected masks from a state-of-the-art segmentation model Segmenter~\cite{strudel2021segmenter} and the groundtruth mask. The results on 512$\times$512 images in Table~\ref{tab3:higher} show that MAT achieves the highest inpainting performance with the mask detected by our method, which is consistent with the performance comparison of corruption detection.

\noindent\textbf{Generalizability on unseen corruption patterns.} To evaluate the generalizability of our model methods across different corruption patterns, we perform training and test over different corruption patterns. Table~\ref{tab3:general} presents the results of our model as well as VCNet by training them on Places dataset while performing testing directly on unseen corruption patterns such as random/constant noise and image content from CelebA-HQ~\cite{karras2018progressive}. It shows that our model consistently outperforms VCNet in all cases. which reveals the better generalizability of our model over VCNet.

\vspace{-2pt}
\subsection{Downstream Image Restoration Tasks}
\label{sec:adaption}
\subsubsection{Blind Image Inpainting}
\vspace{-2pt}


In the experiments of blind image inpainting, we compare our model with VCNet and MPRNet~\cite{zamir2021multi}, a state-of-the-art approach for image restoration. Table~\ref{tab2:blind} presents the inpainting performance on images of 256$\times$256 resolution. Our method achieves the best performance in terms of all metrics and outperforms the other two methods by a large margin. Besides, the qualitative comparison in Figure~\ref{fig10:results} also demonstrates that our model is able to restore higher-quality images than other two methods, benefiting from the precise corruption detection by the proposed hierarchical contrastive learning mechanism as well as the integrated restoration framework. Note that we also provide the user study in the supplementary material.

\begin{figure}[!t]
    \centering
    \begin{minipage}[b]{0.95\linewidth}
    \includegraphics[width=\linewidth]{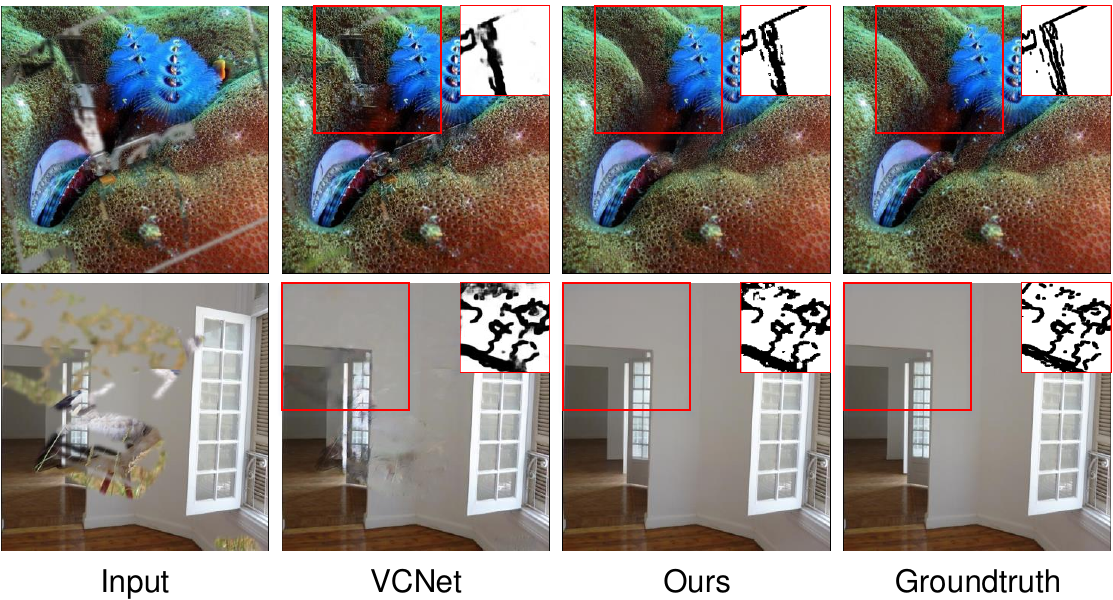}
    \end{minipage}
    \vspace{6pt}
    \caption{Visual results of blind image inpainting on 512$\times$512 resolution images from the Places~\cite{zhou2017places} dataset.}
    \label{fig1:512}
    \vspace{-16pt}
\end{figure}
\begin{table}[!t]
\vspace{-10pt}
\centering
\caption{Blind image inpainting based on detected corruption masks by different methods on $512\times512$ images from Places. MAT~\cite{li2022mat} is finetuned (MAT-F) for fair comparison.}
\resizebox{0.88\linewidth}{!}{
\begin{tabular}{l l | cc|cc}
\toprule
\multicolumn{2}{c|}{Phase} &\multicolumn{2}{c|}{Corruption Detection}&\multicolumn{2}{c}{Image Inpainting}\\
\midrule
\makecell[c]{Corruption Detection}&\makecell[c]{Image Inpainting}&ACC~$\uparrow$&IoU~$\uparrow$&PSNR~$\uparrow$&SSIM~$\uparrow$ \\
\midrule
Groundtruth&MAT-F                           &$-$&$-$&25.34&0.852 \\
Segmenter~\cite{strudel2021segmenter}&MAT-F &0.974&0.962&23.45&0.830 \\
VCNet~\cite{wang2020vcnet}&MAT-F            &0.975&0.963&23.25&0.829 \\
Ours&MAT-F                                  &0.980&0.970&24.53&0.844 \\
\midrule
VCNet&VCNet                                 &0.975&0.963&21.49&0.764 \\
Ours&Ours                                   &0.980&0.970&25.28&0.846 \\
\bottomrule 
\end{tabular}
 }
\label{tab3:higher}
\vspace{-12pt}
\end{table} 
\noindent\textbf{High-resolution blind image inpainting.}
We further conduct experiments on images of 512$\times$512 resolution in Places~\cite{zhou2017places} dataset for blind image inpainting. As shown in Table~\ref{tab3:higher}, our model outperforms VCNet substantially, consistent with the comparison in the case of 256$\times$256 resolution. Besides, the qualitative comparison in Figure~\ref{fig1:512} also validates the superiority of our model. More visual results are presented in the supplementary material. 

\begin{figure}[!t]
    \centering
    \begin{minipage}[b]{0.9\linewidth}
    \includegraphics[width=\linewidth]{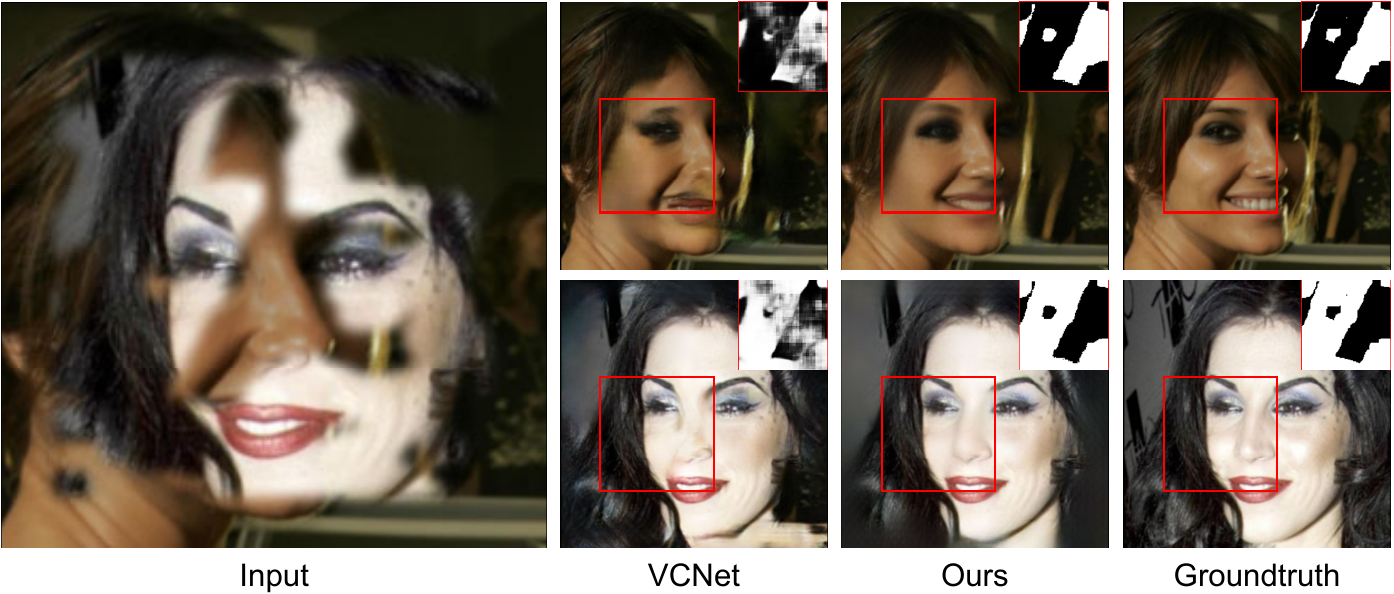}
    \end{minipage}
    \vspace{-8pt}
    \caption{Visual results of bidirectional blind image inpainting on FFHQ~\cite{karras2019style}. More results are shown in the supplementary material.
    \vspace{-12pt}
    }
    \label{fig6:bidirectional}
\end{figure}

\noindent\textbf{Bidirectional blind image inpainting.}
When real natural images are used as the noise content for corruption, we can perform bidirectional image restoration: restore either a complete noise image or a complete background image by reversing the corruption mask. It is quite challenging to perform well in inpainting of both directions. The visual results in Figure~\ref{fig6:bidirectional} show that our model performs more robust than VCNet in both corruption detection and image inpainting.

\begin{figure}[!t]
    \centering
    \begin{minipage}[b]{0.95\linewidth}
    \includegraphics[width=\linewidth]{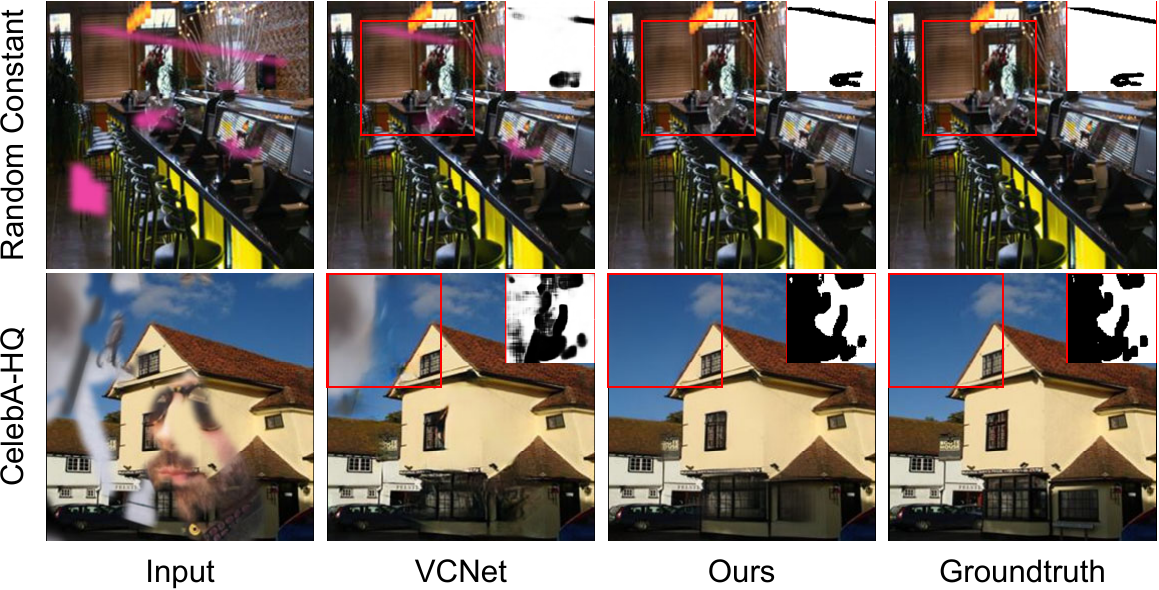}
    \end{minipage}
    \vspace{-4pt}
    \caption{Visual comparison of both corruption detection and blind image inpainting over unseen corruptions. More examples could be found in the supplementary material.
    \vspace{-4pt}
    }
    \label{fig7:rob}
\end{figure}
\begin{table}[!t]
\centering
\vspace{-6pt}
\caption{Performance comparison for blind image inpainting in terms of four evaluation metrics on three benchmark datasets.
}
\resizebox{1\linewidth}{!}{
 \begin{tabular}{c| l| c c c | c c c | c c c }
\toprule
\multicolumn{2}{c|}{Dataset}&\multicolumn{3}{c|}{FFHQ~\cite{karras2019style}} &\multicolumn{3}{c|}{ImageNet~\cite{deng2009imagenet}}&\multicolumn{3}{c}{Places~\cite{zhou2017places}} \\
\cmidrule(lr){1-2} 
\cmidrule(lr){3-5} 
\cmidrule(lr){6-8}
\cmidrule(lr){9-11}
\multicolumn{2}{c|}{Mask ratio (\%)}&0-20 & 20-40 & 40-60 &0-20 & 20-40 & 40-60 &0-20 & 20-40 & 40-60\\
\midrule
\multirow{3}*{PSNR~$\uparrow$}&MPRNet&33.93&26.75&22.06&33.76&26.26&21.50&33.37&25.59&21.08 \\
~&VCNet&29.86&24.51&20.51&29.13&23.39&19.42&29.66&23.19&19.18 \\
~&Ours&\textbf{34.79}&\textbf{27.98}&\textbf{23.54}&\textbf{33.91}&\textbf{26.40}&\textbf{21.80}&\textbf{34.11}&\textbf{26.20}&\textbf{21.66} \\
\midrule
\multirow{3}*{SSIM~$\uparrow$}&MPRNet&0.968&0.898&0.799&0.965&0.882&0.762&0.963&0.872&0.743 \\
~&VCNet&0.933&0.849&0.736&0.924&0.818&0.678&0.942&0.833&0.688 \\
~&Ours&\textbf{0.971}&\textbf{0.910}&\textbf{0.824}&\textbf{0.966}&\textbf{0.884}&\textbf{0.767}&\textbf{0.969}&\textbf{0.887}&\textbf{0.746} \\
\midrule
\multirow{3}*{FID~$\downarrow$}&MPRNet&3.37&12.20&29.74&3.71&17.80&53.38&4.15&17.61&44.73\\
~&VCNet&6.80&14.17&28.90&5.69&21.48&59.11&5.93&18.62&40.21 \\
~&Ours&\textbf{2.37}&\textbf{7.28}&\textbf{15.48}&\textbf{2.69}&\textbf{12.35}&\textbf{41.37}&\textbf{2.88}&\textbf{11.10}&\textbf{31.33} \\
\midrule
\multirow{3}*{LPIPS~$\downarrow$}&MPRNet&0.027&0.096&0.212&0.033&0.123&0.269&0.033&0.131&0.283 \\
~&VCNet&0.044&0.103&0.188&0.045&0.128&0.246&0.046&0.138&0.261 \\
~&Ours&\textbf{0.019}&\textbf{0.065}&\textbf{0.139}&\textbf{0.025}&\textbf{0.098}&\textbf{0.226}&\textbf{0.024}&\textbf{0.095}&\textbf{0.221}\\
\bottomrule
\end{tabular}}
\label{tab2:blind}
\vspace{-8pt}
\end{table} 

\begin{table*}
\begin{floatrow}
\vspace{-2pt}
\begin{minipage}[!t]{0.295\linewidth}
\vspace{0pt}
\centering
\ttabbox{\caption{Generalizability over unseen corruptions for corruption detection. Models are trained on Places.
\vspace{-10pt}}
\label{tab3:general}}
{
\vspace{0pt}
\resizebox{.99\linewidth}{!}{\begin{tabular}{c |l| c c c c}
\toprule
\multicolumn{2}{c|}{Corruption pattern}&\multicolumn{2}{c}{Random constant}&\multicolumn{2}{c}{CelebA-HQ~\cite{karras2018progressive}} \\
\cmidrule(lr){1-2}
\cmidrule(lr){3-4} 
\cmidrule(lr){5-6}
\multicolumn{2}{c|}{Mask ratio (\%)}&0-30&30-60&0-30&30-60\\
\midrule
\multirow{2}*{Acc~$\uparrow$}&VCNet&0.981&0.977&0.975&0.974 \\
~&Ours&\textbf{0.990}&\textbf{0.989}&\textbf{0.986}&\textbf{0.986} \\
\midrule
\multirow{2}*{F1~$\uparrow$}&VCNet&0.988&0.978&0.984&0.976\\
~&Ours&\textbf{0.994}&\textbf{0.990}&\textbf{0.992}&\textbf{0.988} \\
\midrule
\multirow{2}*{BCE~$\downarrow$}&VCNet&\textbf{0.050}&0.078&0.063&0.068 \\
~&Ours&\textbf{0.050}&\textbf{0.049}&\textbf{0.062}&\textbf{0.054} \\
\midrule
\multirow{2}*{IoU~$\uparrow$}&VCNet&0.977&0.960&0.969&0.954 \\
~&Ours&\textbf{0.988}&\textbf{0.981}&\textbf{0.983}&\textbf{0.976} \\
\bottomrule
\end{tabular}}}
\end{minipage}
\hspace{0.5mm}

\begin{minipage}[!t]{0.4\linewidth}
\vspace{4pt}
\centering
\ttabbox{\caption{Generalizability over unseen corruptions for blind inpainting. Models are trained on Places.\vspace{-8pt}}
\label{tab5:general}}{
\resizebox{0.925\linewidth}{!}{
\begin{tabular}{c| l| c c c| c c c}
\toprule
\multicolumn{2}{c|}{Corruption Pattern}&\multicolumn{3}{c|}{Random constant} &\multicolumn{3}{c}{CelebA-HQ~\cite{karras2018progressive}} \\
\cmidrule(lr){1-2} 
\cmidrule(lr){3-5} 
\cmidrule(lr){6-8}
\multicolumn{2}{c|}{Mask ratio (\%)}&0-20 & 20-40 & 40-60 &0-20 & 20-40 & 40-60 \\
\midrule
\multirow{2}*{PSNR~$\uparrow$}&VCNet&30.48&24.33&19.30&29.77&23.21&18.96 \\
~&Ours&\textbf{37.69}&\textbf{29.01}&\textbf{23.05}&\textbf{34.60}&\textbf{26.52}&\textbf{21.82} \\
\midrule
\multirow{2}*{SSIM~$\uparrow$}&VCNet&0.957&0.876&0.732&0.947&0.840&0.685 \\
~&Ours&\textbf{0.985}&\textbf{0.933}&\textbf{0.818}&\textbf{0.972}&\textbf{0.894}&\textbf{0.772} \\
\midrule
\multirow{2}*{FID~$\downarrow$}&VCNet&6.41&16.84&42.54&5.47&17.75&41.19\\
~&Ours&\textbf{1.80}&\textbf{7.45}&\textbf{24.48}&\textbf{2.65}&\textbf{10.50}&\textbf{30.42}\\
\midrule
\multirow{2}*{LPIPS~$\downarrow$}&VCNet&0.051&0.131&0.268&0.041&0.128&0.256 \\
~&Ours&\textbf{0.013}&\textbf{0.063}&\textbf{0.182}&\textbf{0.021}&\textbf{0.089}&\textbf{0.214} \\
\bottomrule
\end{tabular}}}
\end{minipage}

\hspace{0.5mm}
\begin{minipage}[!t]{0.265\linewidth}
\vspace{8pt}
\centering
\ttabbox{\caption{Ablation study on hierarchical interaction mechanism.\vspace{-4pt}}
\label{tab3:ablation}}{
\renewcommand\arraystretch{0.93}
\resizebox{.99\linewidth}{!}{
\begin{tabular}{c |l| c c c c}
\toprule
\multicolumn{2}{c|}{Mask ratio (\%)}&0-30&30-60\\
\midrule
\multirow{3}*{Acc~$\uparrow$}&w/o inter-stage consistency&0.969&0.965\\
~&w/o sample selection&0.975&0.974\\
~&Complete model&\textbf{0.978}&\textbf{0.976}\\
\midrule
\multirow{3}*{F1~$\uparrow$}&w/o inter-stage consistency&0.980&0.969\\
~&w/o sample selection&0.984&0.976\\
~&Complete model&\textbf{0.986}&\textbf{0.978}\\
\midrule
\multirow{3}*{BCE~$\downarrow$}&w/o inter-stage consistency&0.102&0.115\\
~&w/o sample selection&0.094&0.101\\
~&Complete model&\textbf{0.090}&\textbf{0.095}\\
\midrule
\multirow{3}*{IoU~$\uparrow$}&w/o inter-stage consistency&0.962&0.940\\
~&w/o sample selection&0.972&0.957\\
~&Complete model&\textbf{0.975}&\textbf{0.959}\\
\bottomrule
\end{tabular}}}
\end{minipage}

\end{floatrow}
\vspace{-8pt}
\end{table*}

\noindent\textbf{Generalization on unseen corruption patterns.}
\label{sec:general}
Using the same experimental settings as evaluating the generalizability for corruption detection, we also validate the generalizability of our model across unseen corruption patterns on image inpainting. Table~\ref{tab5:general} shows the image inpainting performance of both our model and VCNet on unseen corruption patterns during training. The experimental results indicate that our model performs much better than VCNet. Furthermore, the qualitative comparison in Figure~\ref{fig7:rob} also validates such better generalizability of our model over VCNet on both corruption detection and image inpainting.

\begin{figure}[!t]
    \centering
    \vspace{-12pt}
    \begin{minipage}[b]{1.1\linewidth}
    \includegraphics[width=\linewidth]{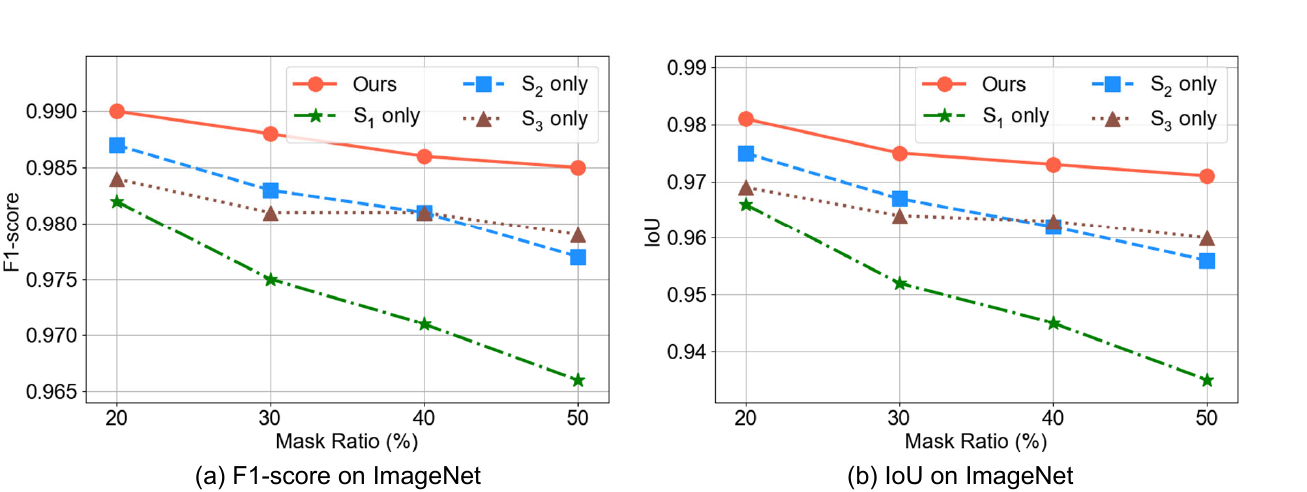}
    \end{minipage}
    \vspace{-10pt}
    \caption{Hierarchical vs single-stage contrastive learning.}
    \vspace{-12pt}
    \label{fig8:hcl}
\end{figure}

\begin{table}[!t]
\centering
\caption{Performance evaluation on watermark removal.}
\vspace{-8pt}
\resizebox{0.575\linewidth}{!}{
\begin{tabular}{l | ccc}
\toprule
Metrics& PSNR$\uparrow$ & SSIM$\uparrow$ & FID$\downarrow$\\
\midrule
SIRF~\cite{zhang2018single}    &34.63&0.978&0.071 \\ 
BS$^2$AM~\cite{cun2020improving}&34.88&0.979&0.028 \\ 
DHAN~\cite{cun2020towards}    &37.67&0.986&0.062 \\ 
BVMR~\cite{hertz2019blind}    &38.28&0.985&0.018 \\ 
Split then Refine~\cite{cun2021split} &41.27&0.991&0.011 \\ 
VCNet~\cite{wang2020vcnet}   &32.66&0.963&0.032 \\ 
Ours       &\textbf{41.88}&\textbf{0.992}&\textbf{0.007} \\ 
\bottomrule
\end{tabular}
 }
 \vspace{-10pt}
\label{tab4:watermark}
\end{table} 

\vspace{-12pt}
\subsubsection{Image Watermark Removal}
\begin{figure}[!t]
    \centering
    \begin{minipage}[b]{1\linewidth}
    \includegraphics[width=\linewidth]{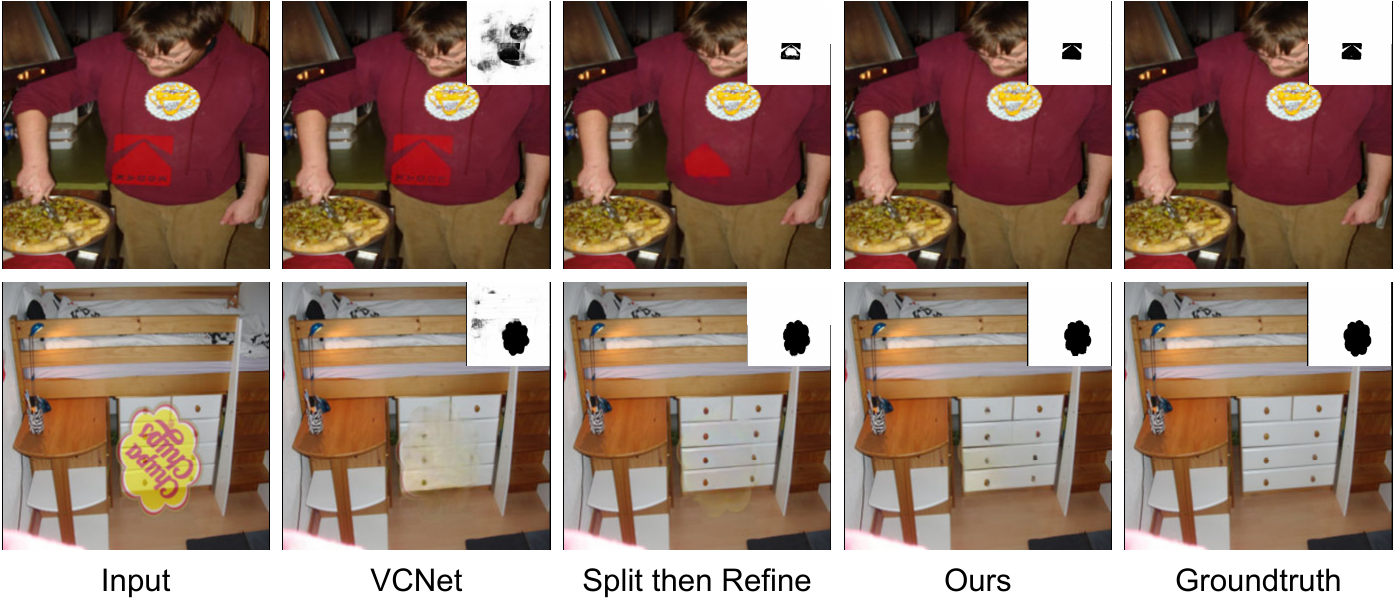}
    \end{minipage}
    \vspace{-20pt}
    \caption{Visual comparison with the state-of-the-art model \emph{Split then Refine}~\cite{cun2021split} and VCNet for watermark removal.
    \vspace{-12pt}
    }
    \label{fig11:down}
\end{figure}

\vspace{-4pt}
In this set of experiments, we further evaluate the performance of our model on LOGO-30K dataset~\cite{cun2021split} for watermark removal. The comparison in Table~\ref{tab4:watermark} shows that our model performs best in terms of all metrics and compares favorably with other specialized methods for watermark removal, which demonstrates the robustness of our model across different image restoration tasks. Moreover, the visual results in Figure~\ref{fig11:down} also validate the superiority of our model over other methods in watermark removal. 

\vspace{-4pt}
\subsection{Ablation Study}
\noindent\textbf{Effect of hierarchical contrastive learning.}
To investigate the effectiveness of the proposed hierarchical contrastive learning framework, we compare its performance with that of single-stage contrastive learning. The results in Figure~\ref{fig8:hcl} demonstrate the distinct advantage of the proposed hierarchical contrastive learning framework.

\noindent\textbf{Effect of hierarchical interaction mechanism.}
We further conduct experiments to investigate the effect of hierarchical interaction mechanism, especially the proposed `inter-stage semantic consistency' and `high-quality sample selection strategy'. Table~\ref{tab3:ablation} presents the concrete experimental results. While both techniques can boost the performance, `inter-stage semantic consistency' is more crucial to the performance due to its essential merit: semantic propagation between stages by feature reusing.

\vspace{-2pt}

\vspace{-2pt}
\section{Conclusion}
\vspace{-4pt}
In this work, we have designed a novel method, namely hierarchical contrastive learning framework, which can automatically detect corruption masks in pixel level and thus allows for blind image corruption restoration without known corruption masks. Extensive quantitative and qualitative comparisons have demonstrated the superior performance of our method over other methods for various corruption restoration tasks and its well generalization ability across different corruption patterns.

\vspace{-4pt}
\section{Acknowledgement}
\vspace{-4pt}
This work was supported in part by the NSFC fund (Grant No. U2013210, 62006060, 62176077), in part by the Shenzhen Key Technical Project under Grant 2022N001, 2020N046, in part by the Guangdong Basic and Applied Basic Research Foundation under Grant (Grant No. 2022A1515010306), in part by Shenzhen Fundamental Research Program (Grant No. JCYJ20210324132210025, JCYJ20220818102415032), and in part by the Guangdong Provincial Key Laboratory of Novel Security Intelligence Technologies (Grant No. 2022B1212010005).

{\small
\bibliographystyle{ieee_fullname}
\bibliography{manuscript}

\begin{thebibliography}{10}\itemsep=-1pt

\bibitem{ba2016layer}
Jimmy~Lei Ba, Jamie~Ryan Kiros, and Geoffrey~E Hinton.
\newblock Layer normalization.
\newblock {\em arXiv preprint arXiv:1607.06450}, 2016.

\bibitem{barnes2009patchmatch}
Connelly Barnes, Eli Shechtman, Adam Finkelstein, and Dan~B Goldman.
\newblock Patchmatch: A randomized correspondence algorithm for structural
  image editing.
\newblock {\em ACM Trans. Graph.}, 28(3):24, 2009.

\bibitem{cai2017blind}
Nian Cai, Zhenghang Su, Zhineng Lin, Han Wang, Zhijing Yang, and Bingo
  Wing-Kuen Ling.
\newblock Blind inpainting using the fully convolutional neural network.
\newblock {\em The Visual Computer}, 33(2):249--261, 2017.

\bibitem{chen2020simple}
Ting Chen, Simon Kornblith, Mohammad Norouzi, and Geoffrey Hinton.
\newblock A simple framework for contrastive learning of visual
  representations.
\newblock In {\em International conference on machine learning}, pages
  1597--1607. PMLR, 2020.

\bibitem{criminisi2004region}
Antonio Criminisi, Patrick P{\'e}rez, and Kentaro Toyama.
\newblock Region filling and object removal by exemplar-based image inpainting.
\newblock {\em IEEE Transactions on image processing}, 13(9):1200--1212, 2004.

\bibitem{cun2020improving}
Xiaodong Cun and Chi-Man Pun.
\newblock Improving the harmony of the composite image by spatial-separated
  attention module.
\newblock {\em IEEE Transactions on Image Processing}, 29:4759--4771, 2020.

\bibitem{cun2021split}
Xiaodong Cun and Chi-Man Pun.
\newblock Split then refine: stacked attention-guided resunets for blind single
  image visible watermark removal.
\newblock In {\em Proceedings of the AAAI Conference on Artificial
  Intelligence}, volume~35, pages 1184--1192, 2021.

\bibitem{cun2020towards}
Xiaodong Cun, Chi-Man Pun, and Cheng Shi.
\newblock Towards ghost-free shadow removal via dual hierarchical aggregation
  network and shadow matting gan.
\newblock In {\em Proceedings of the AAAI Conference on Artificial
  Intelligence}, volume~34, pages 10680--10687, 2020.

\bibitem{deng2009imagenet}
Jia Deng, Wei Dong, Richard Socher, Li-Jia Li, Kai Li, and Li Fei-Fei.
\newblock Imagenet: A large-scale hierarchical image database.
\newblock In {\em 2009 IEEE conference on computer vision and pattern
  recognition}, pages 248--255. Ieee, 2009.

\bibitem{doersch2015unsupervised}
Carl Doersch, Abhinav Gupta, and Alexei~A Efros.
\newblock Unsupervised visual representation learning by context prediction.
\newblock In {\em Proceedings of the IEEE international conference on computer
  vision}, pages 1422--1430, 2015.

\bibitem{dong2022incremental}
Qiaole Dong, Chenjie Cao, and Yanwei Fu.
\newblock Incremental transformer structure enhanced image inpainting with
  masking positional encoding.
\newblock In {\em Proceedings of the IEEE/CVF Conference on Computer Vision and
  Pattern Recognition}, pages 11358--11368, 2022.

\bibitem{efros2001image}
Alexei~A Efros and William~T Freeman.
\newblock Image quilting for texture synthesis and transfer.
\newblock In {\em Proceedings of the 28th Annual Conference on Computer
  Graphics and Interactive Techniques}, pages 341--346, 2001.

\bibitem{feng2022generative}
Xin Feng, Wenjie Pei, Fengjun Li, Fanglin Chen, David Zhang, and Guangming Lu.
\newblock Generative memory-guided semantic reasoning model for image
  inpainting.
\newblock {\em IEEE Transactions on Circuits and Systems for Video Technology},
  2022.

\bibitem{NIPS2014_5ca3e9b1}
Ian Goodfellow, Jean Pouget-Abadie, Mehdi Mirza, Bing Xu, David Warde-Farley,
  Sherjil Ozair, Aaron Courville, and Yoshua Bengio.
\newblock Generative adversarial nets.
\newblock In {\em Advances in Neural Information Processing Systems},
  volume~27. Curran Associates, Inc., 2014.

\bibitem{gulrajani2017improved}
Ishaan Gulrajani, Faruk Ahmed, Martin Arjovsky, Vincent Dumoulin, and Aaron~C
  Courville.
\newblock Improved training of wasserstein gans.
\newblock {\em Advances in neural information processing systems}, 30, 2017.

\bibitem{Guo_2021_ICCV}
Xiefan Guo, Hongyu Yang, and Di Huang.
\newblock Image inpainting via conditional texture and structure dual
  generation.
\newblock In {\em Proceedings of the IEEE/CVF International Conference on
  Computer Vision}, pages 14134--14143, October 2021.

\bibitem{he2020momentum}
Kaiming He, Haoqi Fan, Yuxin Wu, Saining Xie, and Ross Girshick.
\newblock Momentum contrast for unsupervised visual representation learning.
\newblock In {\em Proceedings of the IEEE/CVF conference on computer vision and
  pattern recognition}, pages 9729--9738, 2020.

\bibitem{hendrycks2016gaussian}
Dan Hendrycks and Kevin Gimpel.
\newblock Gaussian error linear units (gelus).
\newblock {\em arXiv preprint arXiv:1606.08415}, 2016.

\bibitem{hertz2019blind}
Amir Hertz, Sharon Fogel, Rana Hanocka, Raja Giryes, and Daniel Cohen-Or.
\newblock Blind visual motif removal from a single image.
\newblock In {\em Proceedings of the IEEE/CVF Conference on Computer Vision and
  Pattern Recognition}, pages 6858--6867, 2019.

\bibitem{jie2020inpainting}
Yong~Shi Jie~Yang, Zhiquan~Qi.
\newblock Learning to incorporate structure knowledge for image inpainting.
\newblock In {\em Proceedings of the AAAI Conference on Artificial
  Intelligence}, volume~34, pages 12605--12612, 2020.

\bibitem{johnson2016perceptual}
Justin Johnson, Alexandre Alahi, and Li Fei-Fei.
\newblock Perceptual losses for real-time style transfer and super-resolution.
\newblock In {\em European Conference on Computer Vision}, pages 694--711.
  Springer, 2016.

\bibitem{karras2018progressive}
Tero Karras, Timo Aila, Samuli Laine, and Jaakko Lehtinen.
\newblock Progressive growing of gans for improved quality, stability, and
  variation.
\newblock In {\em International Conference on Learning Representations}, 2018.

\bibitem{karras2019style}
Tero Karras, Samuli Laine, and Timo Aila.
\newblock A style-based generator architecture for generative adversarial
  networks.
\newblock In {\em Proceedings of the IEEE/CVF conference on computer vision and
  pattern recognition}, pages 4401--4410, 2019.

\bibitem{kingma2014adam}
Diederik~P Kingma and Jimmy Ba.
\newblock Adam: A method for stochastic optimization.
\newblock {\em arXiv preprint arXiv:1412.6980}, 2014.

\bibitem{kingma2013auto}
Diederik~P Kingma and Max Welling.
\newblock Auto-encoding variational bayes.
\newblock {\em arXiv preprint arXiv:1312.6114}, 2013.

\bibitem{Li_2020_CVPR}
Jingyuan Li, Ning Wang, Lefei Zhang, Bo Du, and Dacheng Tao.
\newblock Recurrent feature reasoning for image inpainting.
\newblock In {\em Proceedings of the IEEE/CVF Conference on Computer Vision and
  Pattern Recognition}, June 2020.

\bibitem{li2022mat}
Wenbo Li, Zhe Lin, Kun Zhou, Lu Qi, Yi Wang, and Jiaya Jia.
\newblock Mat: Mask-aware transformer for large hole image inpainting.
\newblock In {\em Proceedings of the IEEE/CVF Conference on Computer Vision and
  Pattern Recognition}, pages 10758--10768, 2022.

\bibitem{li2022misf}
Xiaoguang Li, Qing Guo, Di Lin, Ping Li, Wei Feng, and Song Wnag.
\newblock Misf: Multi-level interactive siamese filtering for high-fidelity
  image inpainting.
\newblock {\em Proceedings of the IEEE/CVF Conference on Computer Vision and
  Pattern Recognition}, 2022.

\bibitem{liu2018image}
Guilin Liu, Fitsum~A Reda, Kevin~J Shih, Ting-Chun Wang, Andrew Tao, and Bryan
  Catanzaro.
\newblock Image inpainting for irregular holes using partial convolutions.
\newblock In {\em Proceedings of the European Conference on Computer Vision},
  pages 85--100, 2018.

\bibitem{Liu2019MEDFE}
Hongyu Liu, Bin Jiang, Yibing Song, Wei Huang, and Chao Yang.
\newblock Rethinking image inpainting via a mutual encoder-decoder with feature
  equalizations.
\newblock In {\em Computer Vision--ECCV 2020: 16th European Conference,
  Glasgow, UK, August 23--28, 2020, Proceedings, Part II 16}, pages 725--741.
  Springer, 2020.

\bibitem{liu2019coherent}
Hongyu Liu, Bin Jiang, Yi Xiao, and Chao Yang.
\newblock Coherent semantic attention for image inpainting.
\newblock In {\em Proceedings of the IEEE/CVF International Conference on
  Computer Vision}, pages 4170--4179, 2019.

\bibitem{liu2022reduce}
Qiankun Liu, Zhentao Tan, Dongdong Chen, Qi Chu, Xiyang Dai, Yinpeng Chen,
  Mengchen Liu, Lu Yuan, and Nenghai Yu.
\newblock Reduce information loss in transformers for pluralistic image
  inpainting.
\newblock In {\em Proceedings of the IEEE/CVF Conference on Computer Vision and
  Pattern Recognition}, pages 11347--11357, 2022.

\bibitem{liu2019deep}
Yang Liu, Jinshan Pan, and Zhixun Su.
\newblock Deep blind image inpainting.
\newblock In {\em International Conference on Intelligent Science and Big Data
  Engineering}, pages 128--141. Springer, 2019.

\bibitem{liu2021swin}
Ze Liu, Yutong Lin, Yue Cao, Han Hu, Yixuan Wei, Zheng Zhang, Stephen Lin, and
  Baining Guo.
\newblock Swin transformer: Hierarchical vision transformer using shifted
  windows.
\newblock In {\em Proceedings of the IEEE/CVF International Conference on
  Computer Vision}, pages 10012--10022, 2021.

\bibitem{Nazeri_2019_ICCV}
Kamyar Nazeri, Eric Ng, Tony Joseph, Faisal Qureshi, and Mehran Ebrahimi.
\newblock Edgeconnect: Structure guided image inpainting using edge prediction.
\newblock In {\em The IEEE International Conference on Computer Vision
  Workshops}, Oct 2019.

\bibitem{simonyan2014very}
Karen Simonyan and Andrew Zisserman.
\newblock Very deep convolutional networks for large-scale image recognition.
\newblock {\em arXiv preprint arXiv:1409.1556}, 2014.

\bibitem{strudel2021segmenter}
Robin Strudel, Ricardo Garcia, Ivan Laptev, and Cordelia Schmid.
\newblock Segmenter: Transformer for semantic segmentation.
\newblock In {\em Proceedings of the IEEE/CVF international conference on
  computer vision}, pages 7262--7272, 2021.

\bibitem{sun2020circle}
Yifan Sun, Changmao Cheng, Yuhan Zhang, Chi Zhang, Liang Zheng, Zhongdao Wang,
  and Yichen Wei.
\newblock Circle loss: A unified perspective of pair similarity optimization.
\newblock In {\em Proceedings of the IEEE/CVF Conference on Computer Vision and
  Pattern Recognition}, pages 6398--6407, 2020.

\bibitem{vaswani2017attention}
Ashish Vaswani, Noam Shazeer, Niki Parmar, Jakob Uszkoreit, Llion Jones,
  Aidan~N Gomez, {\L}ukasz Kaiser, and Illia Polosukhin.
\newblock Attention is all you need.
\newblock {\em Advances in neural information processing systems}, 30, 2017.

\bibitem{wang2021unsupervised}
Longguang Wang, Yingqian Wang, Xiaoyu Dong, Qingyu Xu, Jungang Yang, Wei An,
  and Yulan Guo.
\newblock Unsupervised degradation representation learning for blind
  super-resolution.
\newblock In {\em Proceedings of the IEEE/CVF Conference on Computer Vision and
  Pattern Recognition}, pages 10581--10590, 2021.

\bibitem{wang2021dense}
Xinlong Wang, Rufeng Zhang, Chunhua Shen, Tao Kong, and Lei Li.
\newblock Dense contrastive learning for self-supervised visual pre-training.
\newblock In {\em Proceedings of the IEEE/CVF Conference on Computer Vision and
  Pattern Recognition}, pages 3024--3033, 2021.

\bibitem{wang2020vcnet}
Yi Wang, Ying-Cong Chen, Xin Tao, and Jiaya Jia.
\newblock Vcnet: A robust approach to blind image inpainting.
\newblock In {\em European Conference on Computer Vision}, pages 752--768.
  Springer, 2020.

\bibitem{wu2021contrastive}
Haiyan Wu, Yanyun Qu, Shaohui Lin, Jian Zhou, Ruizhi Qiao, Zhizhong Zhang, Yuan
  Xie, and Lizhuang Ma.
\newblock Contrastive learning for compact single image dehazing.
\newblock In {\em Proceedings of the IEEE/CVF Conference on Computer Vision and
  Pattern Recognition}, pages 10551--10560, 2021.

\bibitem{yu2019free}
Jiahui Yu, Zhe Lin, Jimei Yang, Xiaohui Shen, Xin Lu, and Thomas~S Huang.
\newblock Free-form image inpainting with gated convolution.
\newblock In {\em Proceedings of the IEEE/CVF International Conference on
  Computer Vision}, pages 4471--4480, 2019.

\bibitem{zamir2021multi}
Syed~Waqas Zamir, Aditya Arora, Salman Khan, Munawar Hayat, Fahad~Shahbaz Khan,
  Ming-Hsuan Yang, and Ling Shao.
\newblock Multi-stage progressive image restoration.
\newblock In {\em Proceedings of the IEEE/CVF conference on computer vision and
  pattern recognition}, pages 14821--14831, 2021.

\bibitem{zhang2018single}
Xuaner Zhang, Ren Ng, and Qifeng Chen.
\newblock Single image reflection separation with perceptual losses.
\newblock In {\em Proceedings of the IEEE conference on computer vision and
  pattern recognition}, pages 4786--4794, 2018.

\bibitem{zheng2019pluralistic}
Chuanxia Zheng, Tat-Jen Cham, and Jianfei Cai.
\newblock Pluralistic image completion.
\newblock In {\em Proceedings of the IEEE/CVF Conference on Computer Vision and
  Pattern Recognition}, pages 1438--1447, 2019.

\bibitem{zhou2017places}
Bolei Zhou, Agata Lapedriza, Aditya Khosla, Aude Oliva, and Antonio Torralba.
\newblock Places: A 10 million image database for scene recognition.
\newblock {\em IEEE transactions on pattern analysis and machine intelligence},
  40(6):1452--1464, 2017.

\end{thebibliography}
}

\end{document}